\documentclass{article}

\pdfoutput=1

\PassOptionsToPackage{numbers}{natbib}

\usepackage[final]{neurips_2021}

\usepackage[utf8]{inputenc} 
\usepackage[T1]{fontenc}    
\usepackage{hyperref}       
\usepackage{url}            
\usepackage{booktabs}       
\usepackage{amsfonts}       
\usepackage{nicefrac}       
\usepackage{microtype}      
\usepackage{xcolor}         

\usepackage{graphicx} 
\usepackage{float} 
\usepackage{subfigure} 
\usepackage{multirow}

\title{Deep Learning for Spatiotemporal Modeling of Urbanization}

\author{%
  Tang Li \\
  University of Delaware\\
  Newark, DE 19716 \\
  \texttt{tangli@udel.edu} \\
  \And
  Jing Gao \\
  University of Delaware\\
  Newark, DE 19716 \\
  \texttt{jinggao@udel.edu} \\
  \And
  Xi Peng \\
  University of Delaware\\
  Newark, DE 19716 \\
  \texttt{xipeng@udel.edu} \\
}

\begin{document}

\maketitle

\begin{abstract}
Urbanization has a strong impact on the health and wellbeing of populations across the world. Predictive spatial modeling of urbanization therefore can be a useful tool for effective public health planning. Many spatial urbanization models have been developed using classic machine learning and numerical modeling techniques. However, deep learning with its proven capacity to capture complex spatiotemporal phenomena has not been applied to urbanization modeling. Here we explore the capacity of deep spatial learning for the predictive modeling of urbanization. We treat numerical geospatial data as images with pixels and channels, and enrich the dataset by augmentation, in order to leverage the high capacity of deep learning. Our resulting model can generate end-to-end multi-variable urbanization predictions, and outperforms a state-of-the-art classic machine learning urbanization model in preliminary comparisons.
\end{abstract}
       
\section{Introduction}
\label{intro}
Urbanization is an influential socioeconomic process affecting global population and public health. Predictive spatial modeling of urbanization is an essential tool for related scientific investigation and policy making. Leveraging classic machine learning and numerical modeling techniques, many spatial urbanization models have been developed, including stylized-fact models\cite{jackson2010parameterization}\cite{goldewijk2010k}, developed based on simplifying assumptions about the relationships between different variables, e.g., mapping land development intensity classes by thresholding population density; spatial-interaction-based models\cite{li2017new}\cite{seto2012global}, are commonly utilized in the land cover/land use change studies, e.g., applications of cellular automata\cite{clarke1998loose}\cite{batty2007cities}; data-driven, classic machine learning based models, e.g., Spatially-Explicit Long-term Empirical City developmenT (SELECT) model\cite{gao2019data}, developed by data-driven approaches and newly-available historical time series of fine-spatial-resolution remote sensing observations for making long-term projections of built-up land change\cite{gao2020mapping}. 

Meanwhile, in recent years, deep learning methods have shown an impressive capacity for capturing complex spatiotemporal phenomena, but it has not been applied to urbanization modeling. Here we treat numerical geospatial data as images with pixels and channels and enrich the dataset by augmentation, to leverage deep learning methods, so that our model can generate end-to-end multi-variable urbanization predictions. In preliminary comparisons, our model outperformed a state-of-the-art classic machine learning model (i.e., SELECT) on short-term urbanization projections.

In this paper, we describe our work exploring: (1) the capacity of deep spatial learning for the predictive modeling of urbanization. (2) augmentation methods for the raw data, to enable the deep spatial model training, and (3) comparisons of performance levels of our deep learning model and a classic machine learning model, SELECT, on producing short-term urbanization projections.

\section{Method}
\label{method}
\subsection{Data}
\label{data}
\begin{figure}
\centering
\includegraphics[width=1\textwidth]{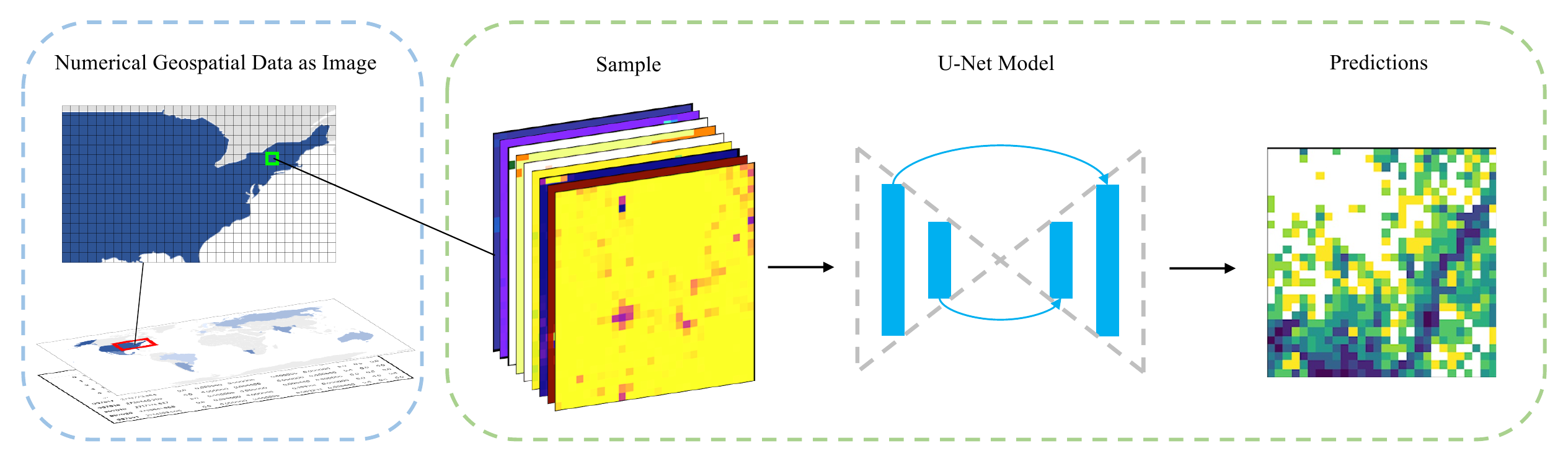}
\caption{Data processing \& modeling}
\label{Fig.1}
\end{figure}

We used an existing dataset for our model development. This dataset was recently compiled and used to develop a classic machine learning based spatial urbanization model, SELECT\cite{gao2019data}. By using this dataset for our model training, we can compare our model’s performance with SELECT’s as a natural benchmark. Figure\ref{Fig.1} illustrates our data processing and modeling method.

In this dataset, the global land area was divided into 997022 grid cells, which means 14km resolution on the equator. We treated each grid cell as a pixel, the whole world as a large image, and attributes (e.g., topographic contexts, population, etc.) as image channels. The input channels are domain experts defined factors that are significantly impacted the urbanization process, the distance to the nearest water body, the distance to the nearest city, the mean elevation, the slope range, the land area within the grid cell/pixel, the population of the year 2000, and the urban land fractions in 1980, 1990, 2000. The target channel stands for the difference of urban land fractions between 2000 and 2010.

We filled water pixels with zeros for each channel, 20-pixel padding was also applied for the sake of cropping. With each land pixel as the center pixel, we densely sampled 997022 image tiles for the whole dataset. The world is divided into 231 national-level regions by the ISO code. We used the pixels in four regions (i.e., the continental U.S., the mainland China, the United Kingdom, and Malawi) as the testing set (138686 pixels, 14\%). The testing set was chosen to include one large developed country/region, one large developing country/region, one small developed country, and one small developing country.

All input images have been cropped into the shape 9x28x28, and the target images have been cropped into the shape 28x28. The center pixel determined the regional belonging of a certain sample. We performed data augmentations by applying vertical flip, horizontal flip, rotation\{$90^{\circ}$, $180^{\circ}$, $270^{\circ}$\} on both input and target image tiles. During training, all water pixels have been masked out from also both the input and target image tiles. 

\subsection{Model}
\label{model}
We treat the urbanization modeling task as a pixel-wise regression task with the inputs and the outputs having the same spatial and temporal resolution. The U-Net model\cite{ronneberger2015u} is a classic deep model for pixel-wise classification, it has been widely used for numerous tasks like biomedical image segmentation\cite{bermudez2018domain}, remote sensing image segmentation\cite{zhu2017deep}, etc. 

We develop a classic U-Net model here because it suits our need to make a pixel-wise prediction. And we made necessary modifications to the model architecture to change it into a pixel-wise regressor. A classic U-Net architecture contains a contracting path to capture the context, and an expanding path to precisely localize\cite{iglovikov2018ternausnet}, without any fully connected layers. The contracting path follows the conventional architecture of a convolutional network with successive convolutions followed by pooling operations, gradually downsampling feature maps, increasing the number of feature maps of each layer simultaneously. The expanding path includes successive steps of upsampling followed by convolution of the feature maps. We used a linear activation function after the last convolution layer instead of logistic activation functions to make regression predictions. Denote the parameter of the model as $\theta$, then the training process of our model can be expressed as:
\begin{equation}
    \hat{\theta} = \arg\min_{\theta} L(Y_i, \hat{Y_i}, M_i)
\end{equation}
where $Y_i$, $\hat{Y_i}$, and $M_i$ are the ground truth image tile matrix, prediction image tile matrix, and mask matrix of the $i$th input image tile, $L$ represents the masked MSE (Mean Squared Error) loss function:
\begin{equation}
    L = \frac{1}{N}\sum_{k=1}^{N} \left(\frac{1}{HW-C}\sum_{i=1}^{H}\sum_{j=1}^{W} m_{ij}(y_{ij} - \hat{y_{ij}})^2 \right)
\end{equation}
where N is the number of samples, H and W are the height and width of the ground truth image tile matrix, C is the number of zeros in the corresponding mask matrix, $m_{ij}$, $y_{ij}$, and $\hat{y_{ij}}$ are the elements on the $i$th row, $j$th column of the mask, ground truth, and prediction matrices.

\section{Experiments}
\label{exp}
\subsection{Urban Land Fraction Change Prediction}
We trained the model on our dataset to make the pixel-wise regression prediction. In the experiments, a classic machine learning model, SELECT\cite{gao2019data}, was used as the baseline model to verify the effectiveness of the proposed method. During the evaluation, the predicted values of duplicate pixels were determined by their median value. Our model and the baseline model were evaluated on four metrics, on both all grid cells and grid cells with observed built-up land fraction > 0 in 2010: (1) Coefficient of Determination\cite{glantz2001primer}\cite{draper1998applied}, $R^2$; (2) Mean Magnitude of Residual; (3) Maximum Magnitude of Residual; and (4) Standard Deviation of Residuals.

\begin{table}[H]
\caption{Summary of residuals evaluated as a whole for short-term estimation (2000–2010) on the change of urban land fractions.}
\label{tbl.1}
\resizebox{\linewidth}{!}{
\begin{tabular}{ccccccccc}
\toprule
\multirow{2}{*}{\begin{tabular}[c]{@{}c@{}}scope: global\end{tabular}} & \multicolumn{4}{c}{All grid cells}                                                                                                                                                                                                                    & \multicolumn{4}{c}{\begin{tabular}[c]{@{}c@{}}Grid cells with observed built-up land fraction \textgreater 0\\ in 2010\end{tabular}}                                                                                                                     \\ \cline{2-9} 
                                                                                       & \begin{tabular}[c]{@{}c@{}}Mean\\ magnitude of\\ residual\end{tabular} & \begin{tabular}[c]{@{}c@{}}Maximum\\ magnitude of\\ residual\end{tabular} & \begin{tabular}[c]{@{}c@{}}Standard\\ deviation of\\ residuals\end{tabular} & $R^2$                 & \begin{tabular}[c]{@{}c@{}}Maximum\\ magnitude of\\ residual\end{tabular} & \begin{tabular}[c]{@{}c@{}}Maximum\\ magnitude of\\ residual\end{tabular} & \begin{tabular}[c]{@{}c@{}}Standard\\ deviation of\\ residuals\end{tabular} & $R^2$                 \\ \midrule
SELECT (baseline)                                                                                 & 0.000367                                                               & 0.435294                                                                  & 0.002041                                                                    & \textgreater{}50\% & 0.000982                                                                  & 0.435294                                                                  & 0.003398                                                                    & \textgreater{}50\% \\
U-Net (sz16)                                                                           & 0.000166                                                               & 0.078438                                                                  & 0.000479                                                                    & 98.739\%           & 0.000457                                                                  & 0.029931                                                                  & 0.000737                                                                    & 98.722\%           \\
U-Net (sz22)                                                                           & 0.000135                                                               & 0.079467                                                                  & 0.000368                                                                    & 99.247\%           & 0.000369                                                                  & 0.025806                                                                  & 0.000551                                                                    & 99.254\%           \\
\textbf{U-Net (sz28)}                                                                  & \textbf{0.000118}                                                      & \textbf{0.025607}                                                         & \textbf{0.000297}                                                           & \textbf{99.498\%}  & \textbf{0.000318}                                                         & \textbf{0.025607}                                                         & \textbf{0.000451}                                                           & \textbf{99.484\%}  \\
\textbf{Multi-task (sz28)}                                                             & \textbf{0.000117}                                                      & \textbf{0.025248}                                                         & \textbf{0.000293}                                                           & \textbf{99.510\%}  & \textbf{0.000314}                                                         & \textbf{0.025248}                                                         & \textbf{0.000445}                                                           & \textbf{99.495\%}  \\ \bottomrule
\end{tabular}}
\end{table}

Table\ref{tbl.1} quantitatively describes the overall residuals of the pixel values. These results showed that in preliminary comparisons, our model outperforms the baseline model on short-term urbanization projections. Our current results show U-Net with augmented data seems a promising method, and grant confidence to continue with further testing. We have also tested different cropping window sizes, and figure out that 28x28 would be a relatively better image size for this specific task. It's reasonable because, for the urbanization prediction, a good receptive field is critical. If the receptive field is too small, there is not enough context has been provided, the model cannot obtain useful information from it; and we cannot make the receptive field too large, because if the physical distance in the real world of the border pixels is too far away from the center, there is no such relationship between their urban expansion process, these distracting information may decline the model's performance.

\subsection{Multi-task Learning}
During urbanization, the change of built-up land fraction and the change of population are a dynamic couple, hence, knowledge sharing between the modeling of these two tasks may increase the performance levels for both target variables. To investigate this possibility, we developed a multi-task modeling experiment. U-Net is known as a typical encoder-decoder architecture model, the contracting path is the encoder part, and the expanding path is the decoder part. Thus we developed our multi-task model architecture by sharing an encoder for both tasks, and provide separate decoders for each of the tasks. Task 1 is the same task of predicting the difference of urban land fractions between 2000 and 2010, task 2 is a new task to predict the difference of populations between the year 2000 and 2010.

\begin{figure}
\centering
\includegraphics[width=0.8\textwidth]{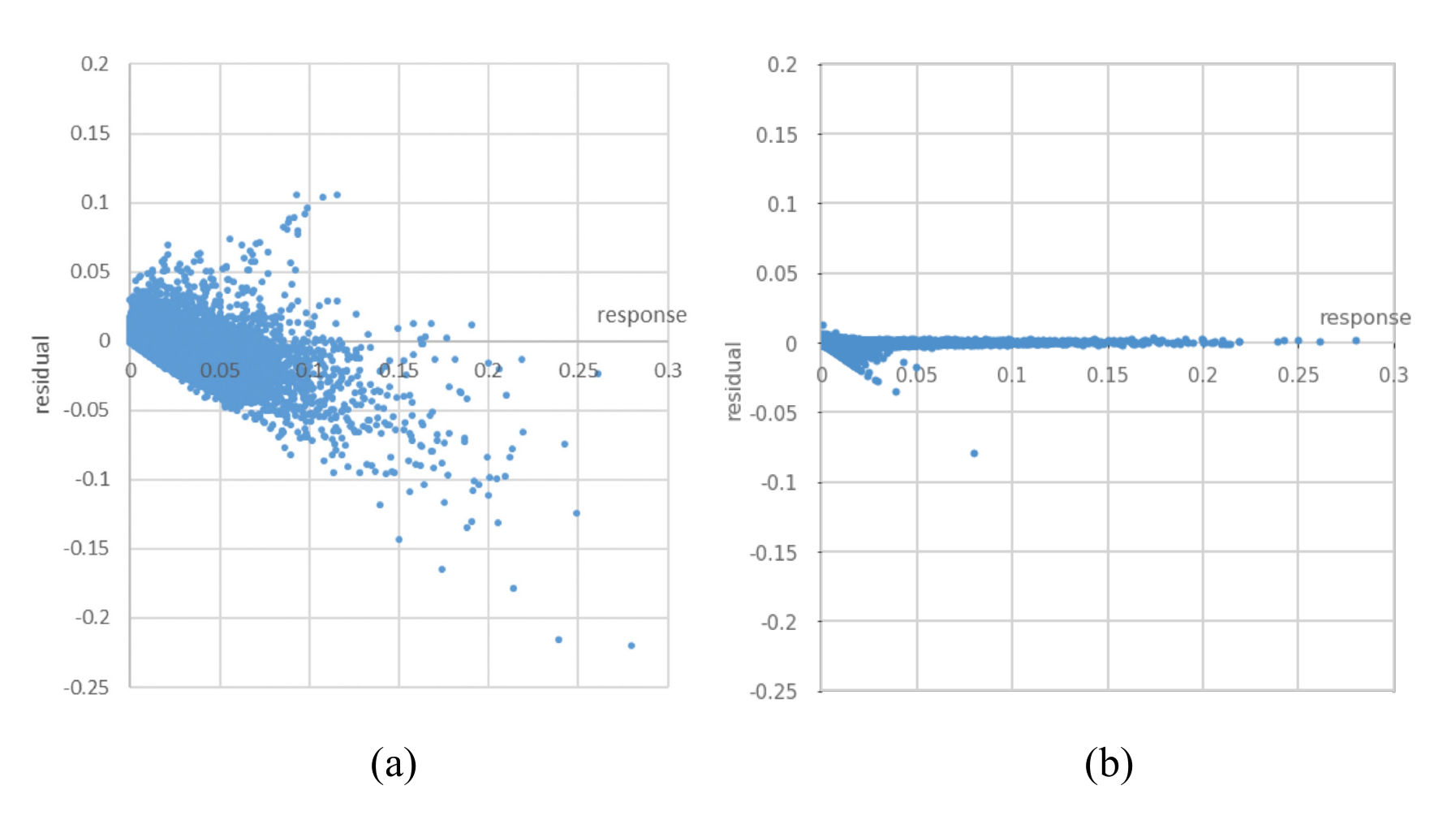}
\caption{Scatterplot of prediction residuals for all global grid cells on the difference of built-up land fractions. (a) is the result of the baseline (SELECT) model\cite{gao2019data}, (b) is the result of our model.}
\label{Fig.2}
\end{figure}

During training, we make use of task 1 pre-trained model as the shared encoder and the task 1 decoder. Freezing the shared encoder and task 1 decoder, train the task 2 decoder from scratch until converge. Then unfreeze the whole model, using a small learning rate to fine-tuning the model until converge. As we can find in Table\ref{tbl.1}, the predicting performance of the difference of urban land fractions between 2000 and 2010 improved slightly, but in Table\ref{tbl.2}, we can see a considerable improvement on task 2. This experiment proved that related tasks will boost the performance of each other, and our method have the capacity to make this kind of modeling possible.

\begin{table}
\caption{Summary of residuals evaluated as a whole for short-term estimation (2000–2010) on the change of populations (normalized).}
\label{tbl.2}
\resizebox{\linewidth}{!}{
\begin{tabular}{ccccccccc}
\toprule
\multirow{2}{*}{\begin{tabular}[c]{@{}c@{}}scope: global\end{tabular}} & \multicolumn{4}{c}{All grid cells}                                                                                                                                                                                                                   & \multicolumn{4}{c}{\begin{tabular}[c]{@{}c@{}}Grid cells with observed built-up land fraction \textgreater 0\\ in 2010\end{tabular}}                                                                                                                    \\ \cline{2-9} 
                                                                                            & \begin{tabular}[c]{@{}c@{}}Mean\\ magnitude of\\ residual\end{tabular} & \begin{tabular}[c]{@{}c@{}}Maximum\\ magnitude of\\ residual\end{tabular} & \begin{tabular}[c]{@{}c@{}}Standard\\ deviation of\\ residuals\end{tabular} & $R^2$                & \begin{tabular}[c]{@{}c@{}}Maximum\\ magnitude of\\ residual\end{tabular} & \begin{tabular}[c]{@{}c@{}}Maximum\\ magnitude of\\ residual\end{tabular} & \begin{tabular}[c]{@{}c@{}}Standard\\ deviation of\\ residuals\end{tabular} & $R^2$                \\ \midrule
Single-task                                                                                 & 0.000136                                                               & 0.018909                                                                  & 0.000210                                                                    & 99.313\%          & 0.000233                                                                  & 0.018909                                                                  & 0.000328                                                                    & 99.403\%          \\
\textbf{Multi-task}                                                                         & \textbf{0.000098}                                                      & \textbf{0.016425}                                                         & \textbf{0.000200}                                                           & \textbf{99.452\%} & \textbf{0.000209}                                                         & \textbf{0.016425}                                                         & \textbf{0.000296}                                                           & \textbf{99.513\%} \\ \bottomrule
\end{tabular}}
\end{table}

\section{Conclusion}
\label{conclusion}
In this paper, we explored the capacity of deep spatial learning methods for predictive urbanization modeling. We augmented the raw data to enrich it and enable the model training. Our model outperformed a classic machine learning model for making short-term urbanization projections. Our results indicate that U-Net with augmented data is a promising method, which provides us with confidence to continue with further testing. For future steps, unboxing what knowledge the deep model has learned might be a valuable direction.

The primary limitation of this work is that we have only tested and evaluated the capability of the deep learning model for making short-term urbanization projections. The model's potential for long-term projections needs further evaluation along with other further testing of the model's behavior. The work has no known negative societal impact. We present here an initial attempt to employ deep spatial learning methods for urbanization modeling, leading the way to eventually provide an advanced tool for public health related scientific investigation and policy making.

\bibliographystyle{unsrt}
\bibliography{main}

\end{document}